\documentclass{article}

    \PassOptionsToPackage{numbers, compress}{natbib}

\usepackage[final]{neurips_2019}

\usepackage[utf8]{inputenc} 
\usepackage[T1]{fontenc}    
\usepackage{url}            
\usepackage{booktabs}       
\usepackage{amsfonts}       
\usepackage{nicefrac}       
\usepackage{microtype}      

\usepackage{color}
\usepackage{tabularx}

\usepackage{amsmath}
\usepackage{amssymb}
\usepackage{mathtools}
\usepackage{graphicx}
\usepackage{float} 
\usepackage{subfigure}
\usepackage{algorithm}  
\usepackage{algpseudocode}  
\usepackage{amsmath}  

\usepackage{comment}
\hypersetup{bookmarksnumbered, colorlinks=true}

\title{Abnormal Client Behavior Detection in Federated Learning}

\author{%
  Suyi Li \thanks{Work done while internship at WeBank.} \\
  HKUST\\
  \texttt{slida@cse.ust.hk} \\
  \And
  Yong Cheng \\
  WeBank \\
  \texttt{petercheng@webank.com} \\
  \And
  Yang Liu \\
  WeBank \\
  \texttt{yangliu@webank.com} \\
  \And
  Wei Wang \\
  HKUST \\
  \texttt{weiwa@cse.ust.hk} \\
  \And
  Tianjian Chen \\
  WeBank \\
  \texttt{tobychen@webank.com}
}

\begin{document}
\maketitle
\begin{abstract}
In federated learning systems, clients are autonomous in that their  behaviors  are not fully  governed    by the server.   Consequently, a client may intentionally or unintentionally deviate  from the prescribed    course of  federated model training, resulting in \textit{abnormal behaviors}, such as turning into a malicious attacker or a malfunctioning client.  Timely detecting those anomalous clients is therefore critical to minimize their adverse impacts.  In this work, we propose to detect anomalous clients at the server side. In particular, we generate low-dimensional surrogates of model weight vectors and use them to perform anomaly detection.  We evaluate our solution through experiments on image classification model training over FEMNIST dataset. Experimental results  show that the proposed  \textit{detection-based} approach significantly outperforms the conventional  \textit{defense-based} methods.   
\end{abstract}

\section{Introduction}\label{sec:introduction}
Federated learning, first proposed by Google in~\cite{McMahan2017,pmlr-v54-mcmahan17a}, is a new paradigm that enables multiple data owners (a.k.a. clients) to \textit{collaboratively} train a machine learning model without sharing their privacy-sensitive data on a server. The federated setting  is a natural fit for  distributed multi-task learning~\cite{Smith2017}. Thanks  to its  potential   for ensuring privacy, federated learning has already found its applications in various fields,  such as mobile internet, finance, insurance, healthcare and smart cities~\cite{QiangYangLiu2019,SystemDesign2019,Intel2018}.

In  a  typical  horizontal federated learning system~\cite{McMahan2017,pmlr-v54-mcmahan17a,QiangYangLiu2019},   there is one server and multiple clients.  Each client performs model training using its own data and transfers local  update to the server for aggregation. The  aggregated model update (a.k.a. global update) is then pulled back from the server by the clients.   This training process repeats iteratively until the model converges or the maximum number of training rounds is reached.  In such a federated setting,   the  federated averaging (\texttt{FedAvg})   algorithm   is  widely-used    for   federated model training,       which takes either the model average  or  the gradient average  of  the  local  model weight or gradient updates from  the  clients~\cite{pmlr-v54-mcmahan17a,yu2019parallel,TianLi2019,XiangyiChen2019,Phong2019}.     



In contrast to its counterpart in distributed machine learning system, the server in a federated learning system has no access to the clients' data,  nor does it have a full control of the clients' behaviors. As a consequence,    a client may deviate from  the normal behaviors during the course  of  federated learning,    which is originally called  \textit{Byzantine attacks} and is referred to as \textit{abnormal client  behavior} in this work.       Abnormal client  behavior  may   be    caused    intentionally,  e.g.,  by  a malicious attacker  disguised as a normal client,  or unintentionally,  e.g.,  by  a client with  hardware  and/or software defects.       It is important to  detect such  abnormal clients,  so as to:  (i) minimize  the impact of anomalous clients;   (ii) report  abnormal clients;    (iii) avoid  model    leakage     to  unintended clients;      and (iv)   prevent  from allocating reward or incentive  to  abnormal   clients~\cite{Incentive2019}. Traditional Byzantine-tolerant algorithms are \textit{defense-based} and sometimes fall short in the federated learning settings with accuracies deteriorated as demonstrated in our experiments. The crux of performance degradation is due to the untargeted defense adopted in existing methods, defending against the attackers at a cost of honest clients. 

In this work, we propose a new approach that leverages a  pre-trained   anomaly detection model to  detect   abnormal client behaviors and eliminate  their  adverse impacts. We employ an  anomaly detection model at the server to run over the local model weight updates received from the clients. We apply \texttt{FedAvg} algorithm~\cite{pmlr-v54-mcmahan17a} to aggregate the model weight updates. Our approach widely applies   to     scenarios  where the   model weight update is sent from a client to the server in various manners, e.g., with no encryption~\cite{pmlr-v54-mcmahan17a}, with additive masking protection~\cite{SecureAggregation2016}, with differential privacy~\cite{DP2018,abadi2016deep}, and in a trusted execution environment (TEE)~\cite{MSTEE2016}.    Since the  model weight  of a deep learning model can easily be  oversized,   we utilize dimensionality reduction   techniques to  generate   surrogates of the  local model weight updates at the server for  anomaly detection.   
Experimental  results show that the proposed  \textit{detection-based} approach outperforms the conventional \textit{defense-based} methods, e.g., achieving an improvement of $10\%$ or more in model accuracy for the jointly trained model.     

\section{Related Work}\label{sec:related_work}
Defending against malicious attackers has been extensively studied in the context of  distributed machine learning,  e.g., popular \textit{defense-based} methods  including \texttt{GeoMed}~\cite{chen2017distributed},  \texttt{Krum}~\cite{NIPS2017_6617}, and  \texttt{Trimmed Mean}~\cite{pmlr-v80-yin18a}.        For  federated learning,  there also  exist several  \textit{defense-based} schemes for minimizing the impact of  malicious attackers, such as the work of~\cite{LipingLi2019}.    The authors of~\cite{Auror2016} proposed a \textit{detection-based} approach for collaborative machine learning.    The work of~\cite{Auror2016} is based on the assumption that  the distribution of the masked features of the training data preserves the distribution of the training data, which is, however, not applicable to federated learning. To our knowledge, this is the first work that applies a \textit{detection-based} approach to the federated learning framework.

\section{Abnormal Client Behavior Detection} \label{sec:adversary_detection}
\subsection{Problem Definition}
We consider a horizontal federated learning system consisting of  one server and $K$ clients that collaboratively train a model   using the \texttt{FedAvg} algorithm~\cite{pmlr-v54-mcmahan17a}.       Due to space limit,  we only present the case where each client sends local model   weight update to the server without encryption,  while leaving  the case with additive masking~\cite{SecureAggregation2016}, differential privacy~\cite{DP2018,abadi2016deep}, and    targeted   model poisoning~\cite{AdversarialLens2019,Backdoor2019} to  the full version of this work.  Our goal  is  to use an anomaly detection model at the server to   detect  the   anomalous clients and to eliminate  their impact on federated model training.

\subsection{The Detection-Based  Approach} 
The  key  idea   of our \textit{detection-based} approach is  that each client  in the federated learning system  is  assigned  a \textit{credit  score},  which is calculated based on the \textit{anomaly score}  produced   by   the anomaly detection model. Assuming $K$ clients participate in federated learning, each client has a number of $n_{k}$ training data points and a local model weight $w_{t+1}^k$ in the $(t+1)$-th global iteration (a.k.a. round). The aggregation in the \texttt{FedAvg} framework~\cite{pmlr-v54-mcmahan17a} is given by 
\begin{equation}\label{eqn:fedavg}
w_{t+1} = \sum_{k=1}^{K} \frac{n_k}{n}w_{t+1}^k,
\end{equation}   

where $w_{t+1}$ represents the global model weight update (i.e., the aggregated model weight update), and $n$ denotes the total number of data points at $K$ clients and we have  $\sum_{k=1}^{K}n_{k} = n$. 


We propose  to replace  term  $\frac{n_k}{n}$ in Eq.~\eqref{eqn:fedavg} with  $\alpha_{t+1}^k$, which may differ in different  rounds:   
\begin{equation}\label{eqn:prop_fedavg}
w_{t+1}    = \sum_{k=1}^{K} {\alpha_{t+1}^k}w_{t+1}^k.
\end{equation}

Given the \textit{anomaly score} $A_{t+1}^{k}$ is assigned to client $k$ in round $t+1$, the \textit{credit score} $\alpha_{t+1}^{k}$ of client $k$ in round $t+1$  is  defined as  
\begin{equation}\label{eqn:alpha_k}
\alpha_{t+1}^k = \frac{  n_{k} \left({A_{t+1}^k}\right)^{-L} } { \sum_{j=1}^K n_{j} \left({A_{t+1}^j}\right)^{-L} }, ~~\forall j = 1, 2, \cdots, K,
\end{equation}

In addition to the fraction of training data points owned by a client, the  \textit{credit score} $\alpha_{t+1}^{k}$ takes the  anomaly score into consideration, leading to targeted defense against abnormal clients. 
The constant  $L$ in Eq.~\eqref{eqn:alpha_k} is a hyperparameter, for tuning  the influence  of $A_{t+1}^{k}$  in  calculating $\alpha_{t+1}^k$. The value of  $L$ should be large if one of the clients  owns a large proportion of data.  Note that, the anomaly score $A_{t+1}^k$ can be calculated by various anomaly detection  models~\cite{chandola2009anomaly}.  In this paper,  we  present an autoencoder-based anomaly detection~\cite{sakurada2014anomaly}, while leaving other anomaly detection schemes to the full version of this work.


Note that since $\sum_{k=1}^{K}\alpha_{t+1}^k = 1$, the convergence of the proposed iterative model averaging procedure in Eq.~\eqref{eqn:prop_fedavg} is guaranteed as long as the convergence of the \texttt{FedAvg} algorithm in Eq.~\eqref{eqn:fedavg} is ensured. This is because it essentially scales down the learning rate.

\subsection{Autoencoder-Based Anomaly Detection}
In the proposed approach, we employ a pre-trained autoencoder model at the server to detect abnormal model weight updates from the clients and hence to detect anomalous clients.   Autoencoder model is known  to be effective for  anomaly detection~\cite{chandola2009anomaly}, especially for high-dimensional data~\cite{sakurada2014anomaly}.     

Denote  $\mathcal{D} = \left\{w_{-1}^1, w_{-1}^2, \cdots w_{-1}^N \right\}$ as the set of model weights  for training the autoencoder model, where subscript $-1$ indicates that  these model weights are accumulated at the server  before the time point for conducting anomaly detection.        An  autoencoder model can then be pre-trained with this dataset $\mathcal{D}$.   In the training process,    a   data point  $w_{-1}^i$ is  firstly compressed into a lower dimensional latent vector by the encoder network and  then reconstructed as $\widetilde{w}_{-1}^i$ by the decoder network~\cite{sakurada2014anomaly}. 
The reconstruction error (a.k.a. mean squared  error (MSE))  of the $i$-th data point is  given by  
\begin{align}
    Err\,( w_{-1}^i) =  {\parallel w_{-1}^i - \widetilde{w}_{-1}^i \parallel}^2.
\end{align}

We can then define anomaly score $A_{t+1}^{k}$ of client $k$ in round $t+1$ as
\begin{equation}\label{eqn:At1k}
A_{t+1}^{k} = \frac{1 + Err\,( w_{t+1}^k)}{ 1 + \sigma_{t+1} },
\end{equation}
where $\sigma_{t+1}$ is defined as  $\sigma_{t+1} =  \min_{j}\left\{Err\,( w_{t+1}^j), ~j = 1, 2, \cdots, K\right\}$.  

For deep learning models, the dimension of the model weight $w_{t+1}^{k}$ can be extremely  large, which  may lead to prohibitive computational complexity for training the autoencoder model and for anomaly detection.    To reduce the computational complexity, we apply dimensionality reduction to the model weight $w_{t+1}^{k}$ to generate low-dimensional surrogates and use them as input to the  autoencoder model. We may apply several existing  dimensionality reduction techniques here~\cite{JMLR:v16:cunningham15a}.    For instance,  when the dimension of the model weight $w_{t+1}^{k}$ is $M$, we may randomly take $\overline{M}$ out of  $M$ elements from $w_{t+1}^{k}$ to form a surrogate vector~\cite{JakubSampling2017}, with $\overline{M}$ being much smaller than  $M$. 
For a specific neural network  model,  we may take the weight of a particular  layer  from the model weight $w_{t+1}^{k}$,  e.g.,  the weights of  the last convolutional layer~\cite{yamada2016weight}.  

After obtaining the anomaly scores of the clients, we can proceed to calculate the credit scores as in Eq.~\eqref{eqn:alpha_k} and run the aggregation operation.  We may further make hard  decisions on which clients are abnormal via \textit{thresholding},  which essentially follows the same procedure as for  binary classification.    
The    credit  score  $\alpha_{t+1}^k$ is  set to  zero  if the anomaly score ${A_{t+1}^k}$  is strictly larger   than  the threshold $A_{t+1}^{th}$; and $\alpha_{t+1}^k$ is  set to  $\frac{n_{k}}{n}$  otherwise.  The  threshold $A_{t+1}^{th}$ can  be chosen as, e.g.,   the average value or the median  of the anomaly scores  $\left\{{A_{t+1}^k}, ~j = 1, 2, \cdots, K\right\}$.    

\section{Experiments}\label{sec:experiments}
\subsection{Experimental Settings}
We evaluate our solution with a $62$-class  image classification task  over  Federated Extended MNIST (FEMNIST) dataset~\cite{caldas2018leaf}.      There are in total $801,074$ samples distributed unevenly among $3,500$  writers in FEMNIST.  Each writer has an average of $229$ samples with standard deviation $89.6$.  Similar to~\cite{TianLi2019,caldas2018leaf}, each  writer in the FEMNIST dataset  represents a client, resulting in a heterogeneous federated setting. During each global round, $K = 20$ writers  will be selected and run mini-batch gradient descent locally with a batch size of $16$ and a learning rate of $0.06$.  Each client performs  local model weight  update  for $20$ epochs. The percentage of abnormal clients is set to be  $30\%$ in every global  round. The anomaly score is defined by equation \ref{eqn:At1k} and the credit score is defined by equation \ref{eqn:alpha_k}. In "Credit Score" approach, we aggregate the clients' model weights by equation \ref{eqn:prop_fedavg}. In "Thresholding" approach, we set the threshold value to be the average value of the anomaly scores.


The image classification  model trained in federated learning  is a CNN model  with two convolutional  layers,  both with $3 \times 3$  kernel,  $32$ channels,  and   $2 \times 2$ max-pooling,   followed by a fully connected layer with $1024$ units,  and a final $62$-dimensional  softmax output layer.   ReLu activation is used in the hidden layers.  This  model has a  similar architecture as  LeNet-5~\cite{lecun1998gradient}.

The autoencoder model used at the server for anomaly detection has four hidden layers.  The input of the autoencoder is a vector with dimensionality of 3000 randomly taken from the last convolutional layer and  the number of  units in each hidden layer is $64$, $32$, $32$, $64$, respectively.   ReLu activation is used in the hidden layers.       The autoencoder model is trained with accumulated local model weight updates at the server, with 
the loss function being reconstruction error (i.e., MSE),     a  batch size of $32$ and a  dropout rate of $0.2$. For calculating credit score,  the hyperparameter  $L$ in Eq.~\eqref{eqn:alpha_k} is set to $2$.  

\subsection{Experimental Results}
Following~\cite{NIPS2017_6617} and~\cite{li2019rsa}, we consider three  adversarial attack models,  namely sign-flipping, additive  noise and gradient ascent. Sign-flipping attack flips the sign of the model weight, and additive noise  attack adds Gaussian noise to  the model weight.  In gradient ascent attack, the  anomalous clients  run gradient ascent locally, instead of gradient descent.   

\begin{figure}[!hb]
\setlength{\abovecaptionskip}{0cm}
\setlength{\belowcaptionskip}{0cm}
\includegraphics[width=\linewidth]{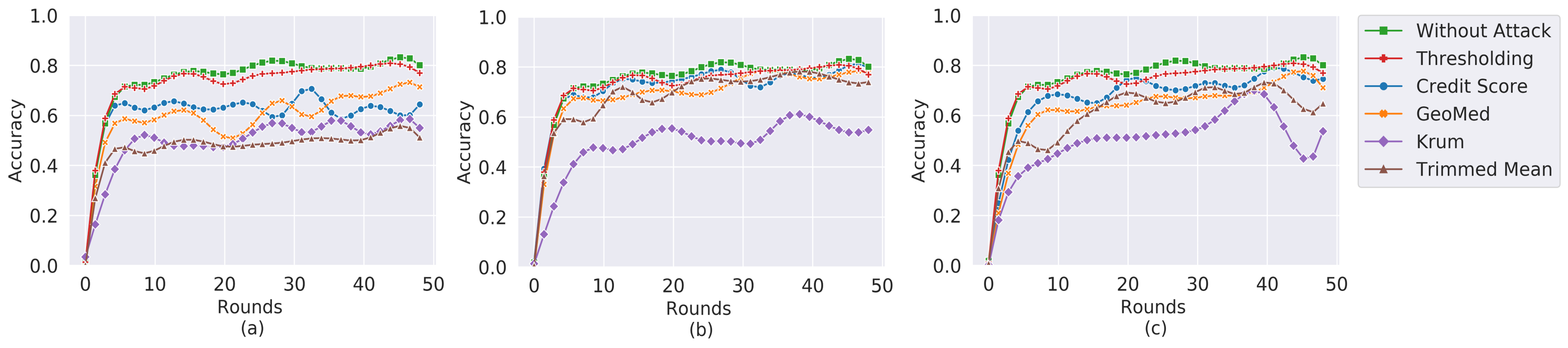}
\caption{Model performance under sign-flipping (a), additive noise (b), gradient ascent (c) attacks}\label{fig:gradientAscent}
\end{figure}

As shown in Fig.~\ref{fig:gradientAscent},  the  proposed  methods ("Thresholding" and "Credit Score")  outperform   the baseline schemes in terms of model accuracy in all settings.    It is interesting to see that, by taking out the abnormal updates (i.e., "Thresholding"),    the proposed \textit{detection-based}   approach achieves almost the same performance as the  \texttt{FedAvg} algorithm without attackers.    While \texttt{GeoMed}~\cite{chen2017distributed} performs the best among the \textit{defense-based} methods,  the proposed \textit{detection-based} approach  has an improvement of $10\%$ or more in model accuracy over \texttt{GeoMed}    under  sign-flipping attacks,   as illustrated in Fig.~\ref{fig:gradientAscent}(a).     Further,  as  discussed in~\cite{NIPS2017_6617},  \texttt{GeoMed}  cannot defend against the large values  sent by abnormal clients.

\texttt{Krum}~\cite{NIPS2017_6617} does  not perform well in defending  against  the considered attacks    because it is not applicable in federated learning,  where data is not  identically and independently distributed (iid).  Only selecting one most appropriate local model weight update among all the updates received from the clients    leads to  high bias in  model aggregation   in non-iid federated settings.  Similarly,   \texttt{Trimmed Mean}~\cite{pmlr-v80-yin18a}  is not effective in defending  against  the  considered attacks   in non-iid federated settings.  Further,   \texttt{Trimmed Mean} and  \texttt{Krum}   require the knowledge of  the fraction of the attackers,   which can not be known  apriori  in federated learning.   




\vspace{-5pt}


\vspace{-5pt}

\section{Summary}\label{sec:summary}
We have shown that autoencoder based anomaly detection can be adopted at the server to detect abnormal local model weight updates from the clients in a federated learning system.  The \textit{detection-based} approach offers the option to opt-out  the anomalous  clients.   Experimental results demonstrate superior performance of the proposed \textit{detection-based} approach over the \textit{defense-based} methods. 

\clearpage
 \addcontentsline{toc}{chapter}{\protect\numberline{}{References}}
 \bibliography{neurips_2019}

\begin{thebibliography}{10}

\bibitem{McMahan2017}
H.~Brendan McMahan, Eider Moore, Daniel Ramage, and Blaise~Aguera y~Arcas.
\newblock Federated learning of deep networks using model averaging.
\newblock {\em arXiv preprint uk.arxiv:1602.05629}, Feb. 2017.

\bibitem{pmlr-v54-mcmahan17a}
Brendan McMahan, Eider Moore, Daniel Ramage, Seth Hampson, and Blaise~Aguera
  y~Arcas.
\newblock Communication-efficient learning of deep networks from decentralized
  data.
\newblock In Aarti Singh and Jerry Zhu, editors, {\em Proceedings of the 20th
  International Conference on Artificial Intelligence and Statistics},
  volume~54 of {\em Proceedings of Machine Learning Research}, pages
  1273--1282, Fort Lauderdale, FL, USA, 20--22 Apr 2017. PMLR.

\bibitem{Smith2017}
Virginia Smith, Chao{-}Kai Chiang, Maziar Sanjabi, and Ameet Talwalkar.
\newblock Federated multi-task learning.
\newblock {\em arXiv preprint arXiv:1705.10467}, Feb. 2018.

\bibitem{QiangYangLiu2019}
Qiang Yang, Yang Liu, Tianjian Chen, and Yongxin Tong.
\newblock Federated machine learning: Concept and applications.
\newblock {\em arXiv preprint arXiv:1902.04885}, Feb. 2019.

\bibitem{SystemDesign2019}
Keith Bonawitz, Hubert Eichner, Wolfgang Grieskamp, Dzmitry Huba, Alex
  Ingerman, Vladimir Ivanov, Chlo{\'{e}} Kiddon, Jakub Konecn{\'{y}}, Stefano
  Mazzocchi, H.~Brendan McMahan, Timon~Van Overveldt, David Petrou, Daniel
  Ramage, and Jason Roselander.
\newblock Towards federated learning at scale: System design.
\newblock {\em arXiv preprint arXiv:1902.01046}, Mar. 2019.

\bibitem{Intel2018}
Micah~J. Sheller, G.~Anthony Reina, Brandon Edwards, Jason Martin, and Spyridon
  Bakas.
\newblock Multi-institutional deep learning modeling without sharing patient
  data: {A} feasibility study on brain tumor segmentation.
\newblock {\em CoRR}, abs/1810.04304, 2018.

\bibitem{yu2019parallel}
Hao Yu, Sen Yang, and Shenghuo Zhu.
\newblock Parallel restarted sgd with faster convergence and less
  communication: Demystifying why model averaging works for deep learning.
\newblock In {\em Proceedings of the AAAI Conference on Artificial
  Intelligence}, volume~33, pages 5693--5700, 2019.

\bibitem{TianLi2019}
Tian Li, Anit~Kumar Sahu, Manzil Zaheer, Maziar Sanjabi, Ameet Talwalkar, and
  Virginia Smith.
\newblock Federated optimization for heterogeneous networks.
\newblock {\em arXiv preprint arXiv:1812.06127}, Jul. 2019.

\bibitem{XiangyiChen2019}
Xiangyi Chen, Tiancong Chen, Haoran Sun, Zhiwei~Steven Wu, and Mingyi Hong.
\newblock Distributed training with heterogeneous data: Bridging median- and
  mean-based algorithms.
\newblock {\em arXiv preprint arXiv:1906.01736}, Jun. 2019.

\bibitem{Phong2019}
L.~T. {Phong} and T.~T. {Phuong}.
\newblock Privacy-preserving deep learning via weight transmission.
\newblock {\em IEEE Transactions on Information Forensics and Security},
  14(11):3003--3015, Apr. 2019.

\bibitem{Incentive2019}
Jiawen Kang, Zehui Xiong, Dusit Niyato, Han Yu, Ying{-}Chang Liang, and Dong~In
  Kim.
\newblock Incentive design for efficient federated learning in mobile networks:
  {A} contract theory approach.
\newblock {\em arXiv preprint arXiv:1905.07479}, May 2019.

\bibitem{SecureAggregation2016}
Keith Bonawitz, Vladimir Ivanov, Ben Kreuter, Antonio Marcedone, H.~Brendan
  McMahan, Sarvar Patel, Daniel Ramage, Aaron Segal, and Karn Seth.
\newblock Practical secure aggregation for federated learning on user-held
  data.
\newblock {\em arXiv preprint arXiv:1611.04482}, Nov. 2016.

\bibitem{DP2018}
Robin~C. Geyer, Tassilo Klein, and Moin Nabi.
\newblock Differentially private federated learning: {A} client level
  perspective.
\newblock {\em arXiv preprint arXiv:1712.07557}, Mar. 2018.

\bibitem{abadi2016deep}
Martin Abadi, Andy Chu, Ian Goodfellow, H~Brendan McMahan, Ilya Mironov, Kunal
  Talwar, and Li~Zhang.
\newblock Deep learning with differential privacy.
\newblock In {\em Proceedings of the 2016 ACM SIGSAC Conference on Computer and
  Communications Security}, pages 308--318. ACM, 2016.

\bibitem{MSTEE2016}
Olga Ohrimenko, Felix Schuster, C{\'e}dric Fournet, Aastha Mehta, Sebastian
  Nowozin, Kapil Vaswani, and Manuel Costa.
\newblock Oblivious multi-party machine learning on trusted processors.
\newblock In {\em Proceedings of the 25th USENIX Conference on Security
  Symposium}, pages 619--636, 2016.

\bibitem{chen2017distributed}
Yudong Chen, Lili Su, and Jiaming Xu.
\newblock Distributed statistical machine learning in adversarial settings:
  Byzantine gradient descent.
\newblock {\em Proceedings of the ACM on Measurement and Analysis of Computing
  Systems}, 1(2):44, 2017.

\bibitem{NIPS2017_6617}
Peva Blanchard, El~Mahdi El~Mhamdi, Rachid Guerraoui, and Julien Stainer.
\newblock Machine learning with adversaries: Byzantine tolerant gradient
  descent.
\newblock In I.~Guyon, U.~V. Luxburg, S.~Bengio, H.~Wallach, R.~Fergus,
  S.~Vishwanathan, and R.~Garnett, editors, {\em Advances in Neural Information
  Processing Systems 30}, pages 119--129. Curran Associates, Inc., 2017.

\bibitem{pmlr-v80-yin18a}
Dong Yin, Yudong Chen, Ramchandran Kannan, and Peter Bartlett.
\newblock {B}yzantine-robust distributed learning: Towards optimal statistical
  rates.
\newblock In Jennifer Dy and Andreas Krause, editors, {\em Proceedings of the
  35th International Conference on Machine Learning}, volume~80 of {\em
  Proceedings of Machine Learning Research}, pages 5650--5659,
  Stockholmsmässan, Stockholm Sweden, 10--15 Jul 2018. PMLR.

\bibitem{LipingLi2019}
Liping Li, Wei Xu, Tianyi Chen, Georgios~B. Giannakis, and Qing Ling.
\newblock {RSA:} byzantine-robust stochastic aggregation methods for
  distributed learning from heterogeneous datasets.
\newblock {\em arXiv preprint arXiv:1811.03761}, Nov. 2018.

\bibitem{Auror2016}
Shiqi Shen, Shruti Tople, and Prateek Saxena.
\newblock {Auror}: {Defending} against poisoning attacks in collaborative deep
  learning systems.
\newblock In {\em Proceedings of the 32Nd Annual Conference on Computer
  Security Applications}, {ACSAC'16}, pages 508--519. ACM, Dec. 2016.

\bibitem{AdversarialLens2019}
Arjun~Nitin Bhagoji, Supriyo Chakraborty, Prateek Mittal, and Seraphin~B. Calo.
\newblock Analyzing federated learning through an adversarial lens.
\newblock {\em arXiv preprint arXiv:1811.12470}, Mar. 2019.

\bibitem{Backdoor2019}
Eugene Bagdasaryan, Andreas Veit, Yiqing Hua, Deborah Estrin, and Vitaly
  Shmatikov.
\newblock How to backdoor federated learning.
\newblock {\em arXiv preprint arXiv:1807.00459}, Aug. 2019.

\bibitem{chandola2009anomaly}
Varun Chandola, Arindam Banerjee, and Vipin Kumar.
\newblock Anomaly detection: {A} survey.
\newblock {\em ACM computing surveys (CSUR)}, 41(3):15, 2009.

\bibitem{sakurada2014anomaly}
Mayu Sakurada and Takehisa Yairi.
\newblock Anomaly detection using autoencoders with nonlinear dimensionality
  reduction.
\newblock In {\em Proceedings of the MLSDA 2014 2nd Workshop on Machine
  Learning for Sensory Data Analysis}, page~4. ACM, 2014.

\bibitem{JMLR:v16:cunningham15a}
John~P. Cunningham and Zoubin Ghahramani.
\newblock Linear dimensionality reduction: Survey, insights, and
  generalizations.
\newblock {\em Journal of Machine Learning Research}, 16(89):2859--2900, 2015.

\bibitem{JakubSampling2017}
Jakub Konecn{\'{y}}, H.~Brendan McMahan, Felix~X. Yu, Peter Richt{\'{a}}rik,
  Ananda~Theertha Suresh, and Dave Bacon.
\newblock Federated learning: Strategies for improving communication
  efficiency.
\newblock {\em arXiv preprint arXiv:1610.05492}, Oct. 2017.

\bibitem{yamada2016weight}
Yasunori Yamada and Tetsuro Morimura.
\newblock Weight features for predicting future model performance of deep
  neural networks.
\newblock In {\em IJCAI}, pages 2231--2237, 2016.

\bibitem{caldas2018leaf}
Sebastian Caldas, Peter Wu, Tian Li, Jakub Kone{\v{c}}n{\`y}, H~Brendan
  McMahan, Virginia Smith, and Ameet Talwalkar.
\newblock Leaf: {A} benchmark for federated settings.
\newblock {\em arXiv preprint arXiv:1812.01097}, 2018.

\bibitem{lecun1998gradient}
Yann LeCun, L{\'e}on Bottou, Yoshua Bengio, Patrick Haffner, et~al.
\newblock Gradient-based learning applied to document recognition.
\newblock {\em Proceedings of the IEEE}, 86(11):2278--2324, 1998.

\bibitem{li2019rsa}
Liping Li, Wei Xu, Tianyi Chen, Georgios~B Giannakis, and Qing Ling.
\newblock Rsa: Byzantine-robust stochastic aggregation methods for distributed
  learning from heterogeneous datasets.
\newblock In {\em Proceedings of the AAAI Conference on Artificial
  Intelligence}, volume~33, pages 1544--1551, 2019.

\end{thebibliography}
 \bibliographystyle{unsrt}

\end{document}